# Predicting Diabetes with Machine Learning Analysis of Income and Health Factors

Fariba Jafari Horestani, M. Mehdi Owrang O*

Department of Computer Science, American University, Washington, DC 20016
Fj9605a@american.edu, owrang@american.edu

# Abstract

In this study, we delve into the intricate relationships between diabetes and a range of health indicators, with a particular focus on the newly added variable of income. Utilizing data from the 2015 Behavioral Risk Factor Surveillance System (BRFSS), we analyze the impact of various factors such as blood pressure, cholesterol, BMI, smoking habits, and more on the prevalence of diabetes. Our comprehensive analysis not only investigates each factor in isolation but also explores their interdependencies and collective influence on diabetes. A novel aspect of our research is the examination of income as a determinant of diabetes risk, which to the best of our knowledge has been relatively underexplored in previous studies. We employ statistical and machine learning techniques to unravel the complex interplay between socio-economic status and diabetes, providing new insights into how financial well-being influences health outcomes. Our research reveals a discernible trend where lower income brackets are associated with a higher incidence of diabetes. In analyzing a blend of 33 variables, including health factors and lifestyle choices, we identified that features such as high blood pressure, high cholesterol, cholesterol checks, income, and Body Mass Index (BMI) are of considerable significance. These elements stand out among the myriad of factors examined, suggesting that they play a pivotal role in the prevalence and management of diabetes.

# 1.Introduction

Diabetes remains a formidable global public health challenge, particularly in the United States where its prevalence is increasing at an alarming rate. According to the CDC's National Diabetes Statistics Report for 2022, cases of diabetes have risen to an estimated 37.3 million, which represents 11.3% of the U.S. population. Among these, 28.7 million have received a diagnosis, while approximately 8.6 million individuals are living with undiagnosed diabetes, indicating a significant gap in disease detection and management. The condition impacts individuals across all social, economic, and ethnic backgrounds, underscoring the complexity of its risk factors and the necessity for widespread, inclusive public health strategies. Furthermore, prediabetes, a serious health condition that increases the risk of developing type 2 diabetes, is present in 26.4 million people aged 65 years or older, which constitutes an astounding 48.8% of the senior population. This data, drawn from the 2015 Behavioral Risk Factor Surveillance System (BRFSS), guides our investigation into the intricate relationships between diabetes and various health indicators. Our aim is to illuminate the multifaceted nature of diabetes risk factors, from biomedical to socio-economic dimensions, to inform more effective public health policies and interventions tailored to these diverse populations.

Historically, research in diabetes has primarily focused on biomedical indicators such as blood pressure, cholesterol, and Body Mass Index (BMI). These factors are well-established in the literature as significant contributors to diabetes risk. For instance, studies have consistently shown a correlation between high BMI and increased diabetes incidence, underscoring the role of obesity as a major risk factor. In recent years, attention has shifted towards understanding how lifestyle choices, such as smoking, diet, and physical activity, interact with traditional biomedical factors in influencing diabetes risk. Behavioral aspects, as explored in several studies, have been found to significantly impact diabetes management and prevention. Moreover, the role of mental and general health in diabetes has gained increasing recognition. Research suggests that mental health conditions, such as depression and anxiety, can exacerbate diabetes symptoms and hinder effective management.

Machine learning has emerged as a vital tool in the predictive modeling of diseases, extending beyond diabetes to various conditions. This is exemplified in our previous study, where we utilized machine learning to forecast the recurrence of breast cancer (Schwarz & Jafari Horestani, 2025). While numerous studies have examined the relationship between lifestyle, health factors, and diabetes, there is a distinct gap in research specifically focusing on the impact of income on diabetes using machine learning. For instance, the study by Chen, Soy, et al., "Using applied machine learning to predict healthcare utilization based on socioeconomic determinants of care," published in the American Journal of Managed Care in 2020, emphasizes the predictive value of social determinants of health (SDH), which encapsulate the conditions in which people live, work, play, and age. However, this study does not isolate income as a direct factor in relation to diabetes.

Our research aims to directly investigate how income levels affect diabetes risk and management, separated from other variables, an area not comprehensively explored in existing literature. By employing advanced machine learning techniques, we intend to delineate the direct correlation between income and diabetes, if any, which has remained uncharted territory until now.

Furthermore, we seek to analyze and recognize the differential impact of lifestyle features such as smoking and alcohol consumption, as well as health factors like cholesterol levels and blood pressure, on diabetes. Understanding which socio-economic factors like income and traditional health and lifestyle indicators have significant influence on diabetes. This approach not only adds a new dimension to diabetes research but also holds the potential to inform more nuanced public health strategies and personalized care protocols.

By integrating a broad spectrum of indicators, from biomedical to socio-economic, our study contributes to a more holistic understanding of diabetes. This comprehensive approach is crucial for informing effective public health policies and individual lifestyle interventions, aiming to improve outcomes for individuals at risk of or living with diabetes.

## 2.Methods

In our study, we employed the Behavioral Risk Factor Surveillance System (BRFSS) dataset, an extensive compilation of health-related data curated by the Centers for Disease Control and Prevention (CDC). The BRFSS, established in 1984, conducts an annual telephone survey that collects data on health-related risk behaviors, chronic health conditions, and the utilization of preventive services. This rich dataset, which includes responses from 441,455 individuals and features 330 distinct variables, provides a comprehensive perspective on public health trends across the United States. For our analysis, we concentrated on a subset of the dataset, specifically the 'diabetes_012_health_indicators_BRFSS2015.csv', which contains 253,680 clean survey responses. The primary variable of interest, 'Diabetes_012', categorizes responses into three classes: '0' for no diabetes or gestational diabetes, '1' for prediabetes, and '2' for diagnosed diabetes, highlighting the presence of class imbalance within the dataset, which comprises 21 feature variables.

To examine the dataset thoroughly, we initially utilized powerful Python libraries like Matplotlib and Seaborn for correlation plots, such as heatmaps, to understand the relationships between variables. After this exploratory analysis, we embarked on a rigorous feature engineering process to identify the most predictive variables. This involved implementing three distinct methods: Lasso regularization to penalize less significant features, Random Forest importance to gauge feature relevance, and Recursive Feature Elimination (RFE) for feature selection efficiency. The convergence of these methods allowed us to isolate high-impact features that were consistently recognized across all models.

before modeling, we'll balance our dataset to address any imbalances between the number of diabetes cases and non-cases. We used the Synthetic Minority Over-sampling Technique (SMOTE). SMOTE works by creating synthetic samples from the minority class (in this case, the number of diabetes cases) instead of creating copies. This helps in balancing the dataset and prevents overfitting that might occur with simple oversampling.

With the key features identified. We used a logistic regression model for the prediction task, which is suitable for binary classification problems. Logistic regression is chosen for its interpretability and efficiency, especially when the relationship between the independent variables and the dependent variable is assumed to be logistic in nature. The model was trained on the balanced and processed training data.

We used different model to predict diabetes based on income. A decision tree classifier was a good choice because it allows us to see how different income levels split the data and contribute to the prediction. Decision trees are also very interpretable, which is valuable when assessing the impact of socioeconomic factors.

We evaluate both models using F1-score, Accuracy, Precision and Recall metrics. After evaluation we used optimization technique to have more accurate result.

Optimization of this model was achieved through the application of GridSearch methodology, enabling us to fine-tune the hyperparameters effectively. Given our specific interest in the influence of income on diabetes prevalence, we channeled the income data into a decision tree model to ascertain its predictive capacity for diabetes.

Throughout the study, we utilized Python programming language for all coding, data analysis, and visualization tasks, ensuring a robust and reproducible research process.

## 3.Result and Discussion

## 3.1. Dataset Visualization

To lay the foundation for a nuanced analysis, we employed a range of visualization techniques aimed at uncovering underlying patterns and relationships within the BRFSS2015 data. Utilizing powerful Python libraries like Matplotlib and Seaborn, we constructed a series of graphs that bring to light the intricate dynamics between demographic variables, health assessments, and income levels.

### 3.1.1. Correlation Heatmap Analysis

A correlation heatmap was utilized to identify relationships between health indicators and socioeconomic variables within our dataset. As figure 1 is showing, a strong positive correlation between 'General Health' (GenHlth) and 'Physical Health' (PhysHlth) suggests that individuals who rate their overall health poorly also report more physically unhealthy days. An inverse correlation between 'General Health' (GenHlth) and 'Income' indicates that higher income levels are associated with better self-rated health.

These visual correlations, while not indicative of causation, highlight significant trends that warrant further statistical exploration. The heatmap's color gradations—from deep reds to light yellows—intuitively signal the strength of these relationships, guiding our analysis towards the socioeconomic factors that may influence health outcomes.

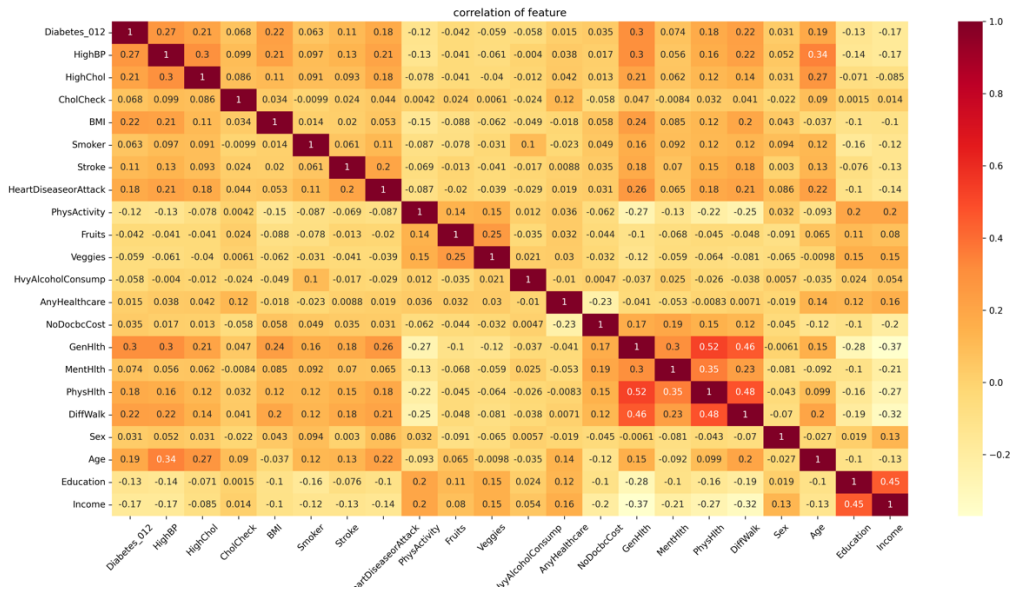

**Figure 1**. Correction Heatmap

### 3.1.2. Visualization income in the dataset

This bar chart in figure 2 presents the distribution of income categories within our dataset, which is a pivotal variable in our analysis of the impact of income on diabetes prevalence. Income levels are divided into eight distinct categories, ranging from an annual income of less than $10,000 to an annual income exceeding $75,000. As depicted in the chart, there is a noticeable concentration of responses in the highest income category, indicating that a substantial portion of the dataset's participants report an annual income greater than $75,000. This distribution provides crucial context for our study as it highlights the income spectrum of the participants, thereby enabling us to assess the influence of income on diabetes within a diverse economic demographic.

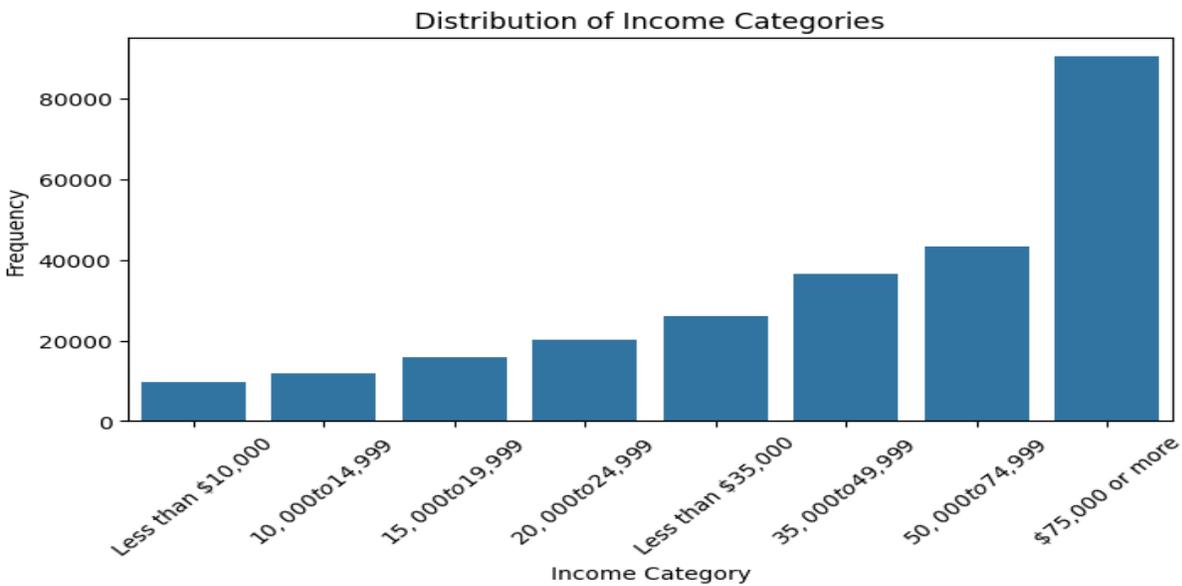

**Figure 2**. Distribution of income

## 3.2. Feature Engineering

In the development of our predictive model, we have employed a multi-faceted approach to feature engineering to ensure that we utilize the most significant predictors in our dataset. This process involved the use of three distinct techniques: Lasso regularization, Random Forest importance, and Recursive Feature Elimination (RFE), each contributing unique insights due to their methodological differences.

### 3.2.1. L1

Lasso, or L1 regularization, was applied to penalize the absolute size of the coefficients in our model. By imposing this constraint, we effectively shrunk less important feature coefficients to zero, thus performing feature selection as figure 3 is showing. This method is particularly useful when we have a high number of features, as it helps in reducing the model complexity and preventing overfitting. The output from the Lasso model provided us with a visual bar plot, from which we could discern the features with non-zero coefficients, signifying their importance.

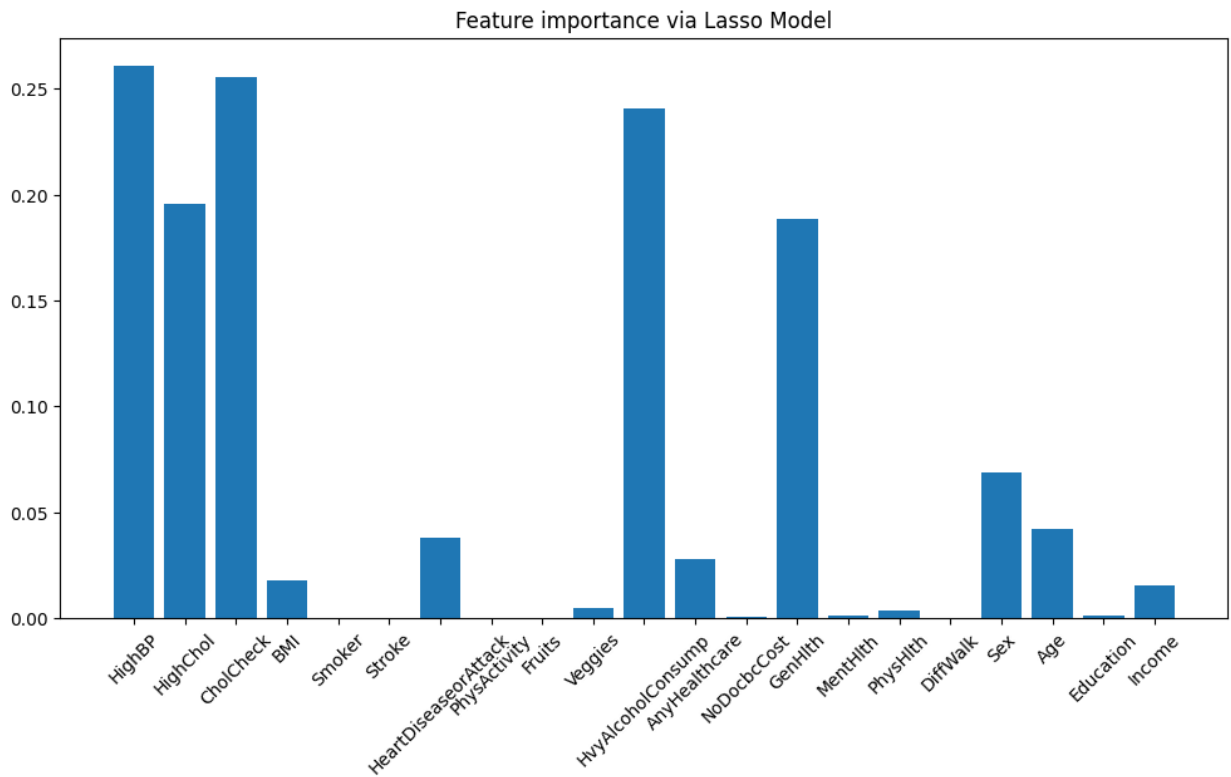

**Figure 3**. Determining Feature Importance Using the Lasso Model

### 3.2.2. Random Forest

The Random Forest algorithm offers a measure of features importance which are computed from the aggregated contribution of each feature in reducing the variance within the numerous decision trees constructed during the modeling process. This non-linear approach is adept at capturing complex interactions between features and is robust to outliers. According to figure 4 the feature importance plot from the Random Forest model complements our analysis by highlighting features that significantly contribute to the prediction accuracy.

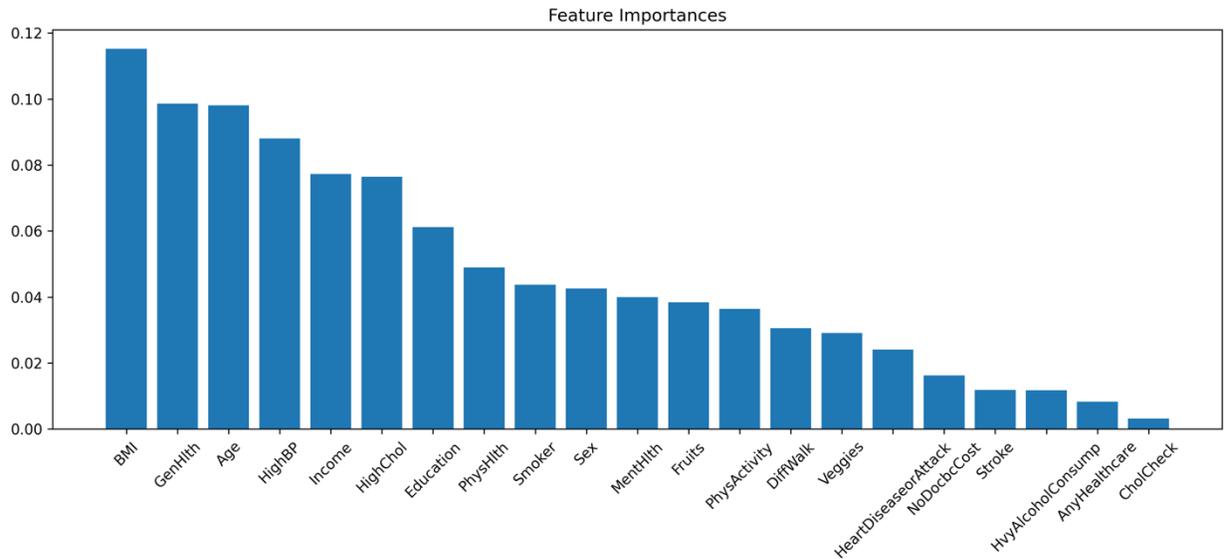

**Figure 4**. Determining Feature Importance Using the Random Forest Model

### 3.2.3. Recursive Feature Elimination (RFE)

RFE is a wrapper-type feature selection method that works by recursively removing the least important features based on the model performance. By training the model multiple times and eliminating features, RFE helps in pinpointing the set of features that are most critical. The RFE output lists features along with a Boolean indicator for selection and a rank indicating the order of elimination as figure 5 is indicating this result.

```
    n_iter_i = _check_optimize_result(
Column: HighBP, Selected True, Rank: 1
Column: HighChol, Selected True, Rank: 1
Column: CholCheck, Selected True, Rank: 1
Column: BMI, Selected False, Rank: 10
Column: Smoker, Selected False, Rank: 6
Column: Stroke, Selected True, Rank: 1
Column: HeartDiseaseorAttack, Selected True, Rank: 1
Column: PhysActivity, Selected False, Rank: 5
Column: Fruits, Selected False, Rank: 8
Column: Veggies, Selected False, Rank: 9
Column: HvyAlcoholConsump, Selected True, Rank: 1
Column: AnyHealthcare, Selected True, Rank: 1
Column: NoDocbcCost, Selected True, Rank: 1
Column: GenHlth, Selected True, Rank: 1
Column: MentHlth, Selected False, Rank: 11
Column: PhysHlth, Selected False, Rank: 12
Column: DiffWalk, Selected False, Rank: 2
Column: Sex, Selected True, Rank: 1
Column: Age, Selected False, Rank: 4
Column: Education, Selected False, Rank: 3
Column: Income, Selected False, Rank: 7
/usr/local/lib/python3.10/dist-packages/sklearn/linear_model/_logistic.py:458: ConvergenceWarning: lbfgs failed to converge (status=1):
STOP: TOTAL NO. of ITERATIONS REACHED LIMIT.
```

**Figure 5**. Determining Feature Importance Using the RFE

### 3.2.4. Interpreting the result of feature selection techniques

By integrating these three methods, we aimed to balance the strengths and weaknesses inherent to each. For instance, while Lasso is linear and may not capture complex patterns, Random Forest can, but it might be computationally intensive. RFE, while potentially prone to overfitting, provides another layer of validation to our feature selection process. The combined use of these methods allowed us to triangulate the feature importance, increasing our confidence in the selected features. For the Lasso and Random Forest importance plots, we can visually inspect which features have the tallest bars (indicating higher importance). Combining this information with the RFE results, we can get an overall sense of feature importance.

Features like HighBP, HighChol, and CholCheck appear to be important across all three methods, as they are selected by RFE and show significant importance in the Lasso and Random Forest plots corroborated by literature that acknowledges their impact on diabetes (American Heart Association). Other health-related features such as Stroke, HeartDiseaseorAttack, GenHlth, and BMI seem to have varying levels of importance but are generally considered influential across the methods being notably documented in population health (diaTribe Foundation). Demographic features such as Age, Sex, Education, and Income show importance in the Random Forest and Lasso results, although their ranks may vary. Lifestyle-related features such as Smoker, PhysActivity, Fruits, Veggies, and HvyAlcoholConsump also appear to be relevant, with Smoker and HvyAlcoholConsump being highlighted by RFE and showing up in the plots.

## 3.3 Predicting Diabetes Prevalence Using Key Health Indicators

In this section, we will develop a predictive model to estimate the likelihood of diabetes occurrence based on critical health indicators: High Blood Pressure (HighBP), High Cholesterol (HighChol), Cholesterol Check (CholCheck), Smoking Status (Smoker), Heavy Alcohol Consumption (HvyAlcoholConsump), and Body Mass Index (BMI). The choice of these features is informed by their strong association with diabetes as identified through our feature engineering process.

Before balancing dataset, we train the regression model and its accuracy on the test set was approximately 84.8%. But the Percision and recall for class 1 (prediabetes) were both, indicating the model is not predicting this class correctly. This is likely due to class imbalance in the dataset (with more non-diabetic than diabetic instances). Therefore, before modeling, we'll balance our dataset to address any imbalances between the number of diabetes cases and non-cases. A common technique for balancing datasets is Synthetic Minority Over-sampling Technique (SMOTE). SMOTE works by creating synthetic samples from the minority class (in this case, the number of diabetes cases) instead of creating copies. This helps in balancing the dataset and prevents overfitting that might occur with simple oversampling.

After initial training with balanced dataset the Health Indicators Logistic Regression Model yielded the following performance metrics that figure 6 is showing that.

Precision: The model had a precision of 0.71 for classifying non-diabetic individuals and 0.70 for diabetic individuals. This indicates that when the model predicted an individual's status (diabetic or non-diabetic), it was correct approximately 70-71% of the time.

Recall: The recall was 0.69 for non-diabetic predictions and 0.72 for diabetic predictions. This means that the model correctly identified 69% of the actual non-diabetic individuals and 72% of the actual diabetic individuals.

F1-Score: The F1-score, which is a balance between precision and recall, was 0.70 for non-diabetic and 0.71 for diabetic individuals, showing a balanced precision-recall trade-off.

Accuracy: The overall accuracy of the model was 0.71, suggesting that it correctly identified 71% of all cases.

AUC Score: The AUC score was 0.7051, indicating that the model had a 70.51% chance of distinguishing between a diabetic and non-diabetic individual.

```
Health Indicators Model Report:
              precision    recall  f1-score   support

           0       0.71      0.69      0.70     42916
           1       0.70      0.72      0.71     42566

    accuracy                           0.71     85482
   macro avg       0.71      0.71      0.71     85482
weighted avg       0.71      0.71      0.71     85482

Health Indicators Model AUC Score: 0.7051436982236732
```

**Figure 6**. performance metrics of Regression Model before optimization

After establishing baseline performances for our Health Indicators Models, we embarked on a rigorous hyperparameter tuning process. This optimization aimed to refine the models' configurations to improve their predictive accuracy.

Through grid search cross-validation, we identified the optimal hyperparameters as 'C': 1 and 'solver': 'liblinear'. The tuning process, involving 50 fits across 5 folds, led to a marked improvement in the model's AUC score, elevating it to 0.7743. This optimization solidifies the model's capability to predict diabetes more accurately by leveraging health-related features.

After optimization, the visualizations indicate an enhanced model performance. In figure 7 the ROC curve, which plots the true positive rate against the false positive rate, shows a larger area under the curve (AUC = 0.77). This suggests that the model's ability to differentiate between diabetic and non-diabetic individuals has improved after optimization. The increase in the AUC score indicates a higher likelihood of correctly classifying individuals.

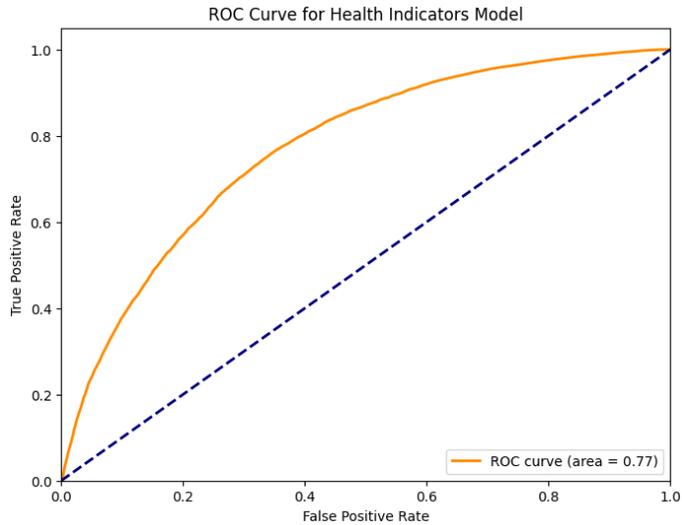

**Figure 7**. ROC Cure, Regression Model after optimization

Figure 8 Precision-Recall Curve appears closer to the top-right corner, indicating an improvement in both precision (the model's correctness when predicting diabetes) and recall (the model's ability to find all the diabetic cases).

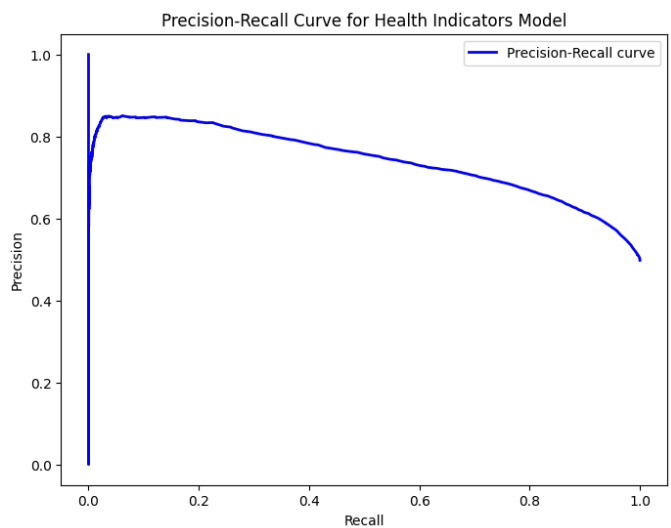

**Figure 8**. Precision-recall curve, Regression Model after optimization

Figure 9 the confusion matrix shows the breakdown of true positives, false positives, true negatives, and false negatives. After optimization, we would expect a higher number of true positives and true negatives (top-left and bottom-right of the matrix, respectively), indicating better performance.

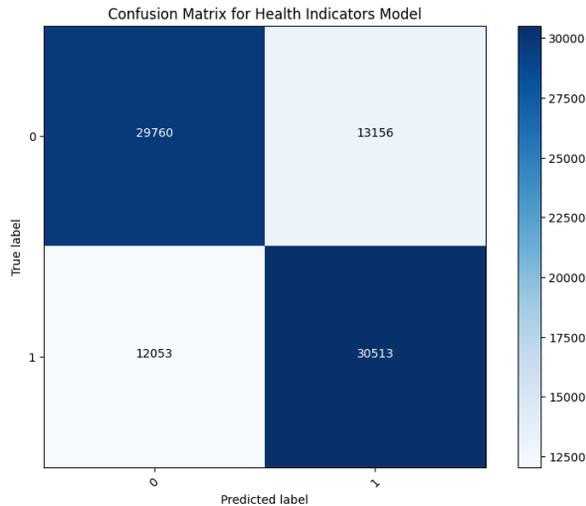

**Figure 9**. Confusion Matrix, Regression Model after optimization

The results post-optimization not only demonstrate the potential of machine learning in medical predictive analytics but also highlight the need for a nuanced approach to model development. The selection of an appropriate model, along with meticulous feature selection and hyperparameter tuning, is paramount to developing tools that can effectively inform healthcare decisions.

### 3.4. Assessing the Impact of Income Status on Diabetes Risk

In this analysis, we examined how income levels alone can predict diabetes. Given the nature of the data, we used a decision tree classifier (DCT) that is a good choice because it allows us to see how different income levels split the data and contribute to the prediction. The initial performance of the Income Decision Tree Model provided a foundational understanding of how well income alone can predict diabetes. The DCT model yielded the following performance metrics that figure 10 is showing that.

Precision: The precision for non-diabetic predictions was 0.61, and for diabetic predictions, it was 0.59. This indicates that the model's predictions were correct about 59-61% of the time when asserting an individual's diabetic status based on income alone.

Recall: The recall for the model was 0.56 for non-diabetic and 0.64 for diabetic classifications. This metric signifies that the model identified 56% of all actual non-diabetic individuals and 64% of all actual diabetic individuals in the dataset.

F1-Score: The F1-score was 0.58 for non-diabetic and 0.62 for diabetic predictions, reflecting a moderate balance between precision and recall, skewed slightly towards better identification of diabetic individuals.

Accuracy: Overall accuracy stood at 0.60, meaning that 60% of all classifications (diabetic or non-diabetic) were correct.

AUC Score: The Area Under the Curve was 0.601, implying that the model had a 60.1% chance of correctly distinguishing between a diabetic and a non-diabetic individual based solely on income data.

```
Health Indicators Model AUC Score: 0.7051436982236732

Income Model Report:
              precision    recall  f1-score   support

           0       0.61      0.56      0.58     42916
           1       0.59      0.64      0.62     42566

    accuracy                           0.60     85482
   macro avg       0.60      0.60      0.60     85482
weighted avg       0.60      0.60      0.60     85482

Income Model AUC Score: 0.6010309892201988
```

**Figure 10**. performance metrics, DCT Model before optimization

According to the evaluation result, the Decision Tree Model's initial predictive power was found to be moderate, with an Area Under the Receiver Operating Characteristic (ROC) Curve (AUC) score of 0.601. To improve this, we embarked on a hyperparameter tuning exercise, seeking to enhance the model's ability to discern between individuals with and without diabetes based solely on their income level. Hyperparameters are the configurable settings used to steer the learning process and dictate the model's architecture. For decision trees, key hyperparameters include:

max_depth: Controls the maximum depth (or length) of the tree. Deeper trees can model more complex patterns but are also more prone to overfitting.

min_samples_split: Determines the minimum number of samples required to split an internal node. Higher values prevent creating nodes that only fit a small number of observations.

min_samples_leaf: The minimum number of samples required to be at a leaf node. Setting this parameter helps in smoothing the model, especially for regression tasks.

We utilized GridSearchCV, an exhaustive search over specified hyperparameter values for an estimator, to optimize these parameters. The optimization process revealed that the Decision Tree Model achieved the best performance at the default settings of max_depth (none), indicating that the model benefits from deeper trees that can capture more complex patterns in the data. Additionally, the min_samples_leaf was set to 1 and min_samples_split to 2, both of which are the default settings for a decision tree and suggest that the model performs best with the flexibility to create leaves and splits without restriction.

Following the optimization process, the visual and numerical results indicate the following improvements:

The ROC curve shows an increased area under the curve (AUC = 0.63), denoting an enhanced capability of the model to discriminate between the diabetic and non-diabetic individuals after optimization. The growth in AUC score is an indicator of a model that is becoming more effective in classification tasks.

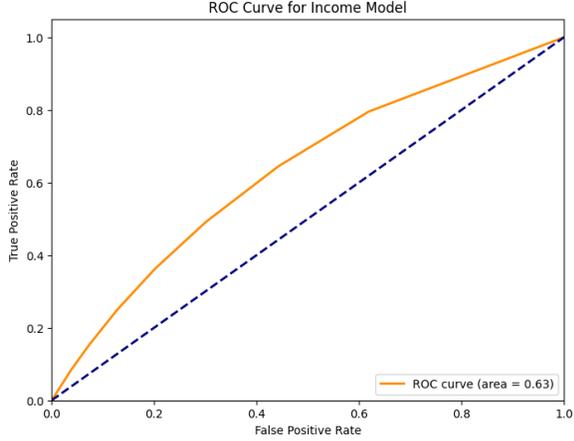

**Figure 11**. ROC Cure, DCT Model after optimization

In the figure 11, the Precision-Recall curve post-optimization suggests some improvement in the model's ability to balance precision and recall, but it also highlights the challenges in achieving high performance when using income as the sole predictor.

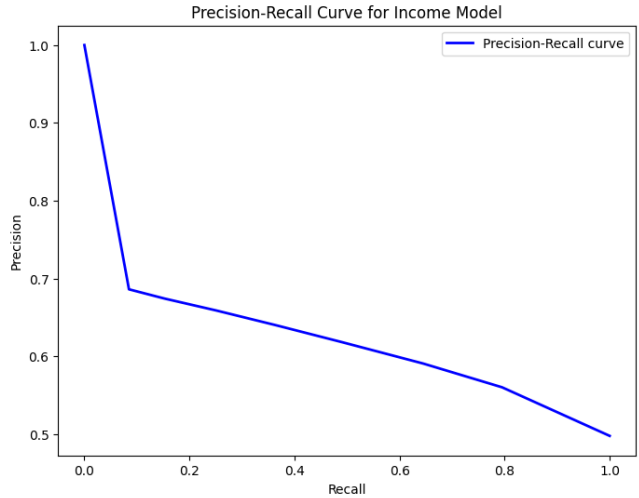

**Figure 12**. Precision-recall curve, DCT Model after optimization

A post-optimization confusion matrix, figure 12, would ideally show more true positives and true negatives, indicative of a better-performing model. However, given the complexity of predicting diabetes from income alone, the improvements might be marginal.

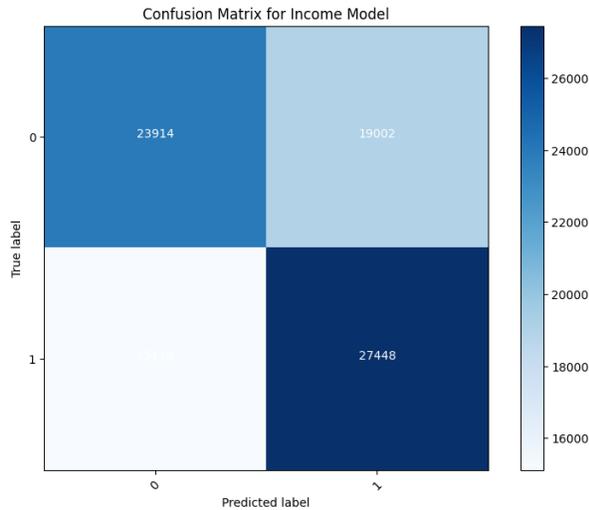

**Figure 12**. Confusion Matrix, Regr after optimization

Our logistic regression model, optimized for health-related features, demonstrated a notable capacity to predict diabetes with an AUC score improving to 0.77 post-optimization. This model underscored the profound influence of physiological factors such as blood pressure, cholesterol, and BMI in forecasting diabetes. The optimization process, particularly the fine-tuning of the regularization strength, was instrumental in enhancing the model's predictive accuracy. The refined model presents a robust tool that could potentially aid in early diabetes detection and management, provided it is used within the context of comprehensive clinical assessment. Conversely, the decision tree model based on income offered a window into the socioeconomic aspects of diabetes risk. Despite a rigorous hyperparameter tuning process, the model's predictive power saw only a modest increase in AUC to 0.633. This increment, while insightful, brought to light the limitations of using income as a sole predictor, considering the multifaceted nature of diabetes etiology. It became evident that socioeconomic status, while reflective of certain risk factors for diabetes, cannot singularly account for the complex interplay of lifestyle, genetic, and environmental elements.

The synthesis of findings from both models paints a nuanced picture of diabetes risk. Health indicators, with their direct physiological relevance, exhibit a stronger predictive relationship with diabetes than income. However, the role of socioeconomic status, as reflected in the income model, cannot be discounted. It hints at a gradient of diabetes risk across different income levels, suggesting that lower socioeconomic status may be associated with higher diabetes prevalence. The discrepancy in the predictive performance of the two models accentuates the necessity for a multifactorial approach to diabetes risk assessment. A comprehensive model that integrates both health and socioeconomic factors may offer a more accurate and holistic understanding of diabetes risk.

In future work, the integration of the logistic regression model with other data sources, such as electronic health records or genetic information, could potentially unveil more profound insights into diabetes risk. Moreover, exploring machine learning methodologies that transcend the limitations of logistic regression, such as ensemble methods or deep learning, may offer further advances in the predictive accuracy and clinical utility of such models.

## 4.Conclusion

In conclusion, this study has successfully extended the body of research into diabetes by incorporating a comprehensive set of variables, including the often-overlooked factor of income. Our findings highlight a clear association between lower income levels and an increased risk of diabetes, echoing the broader understanding that socio-economic factors are crucial determinants of health. The use of the BRFSS 2015 dataset has allowed us to delve deeply into the complex interplay of health indicators and lifestyle choices, emphasizing the significance of high blood pressure, cholesterol levels, and BMI in diabetes prevalence. By integrating sophisticated statistical and machine learning tools, our analysis not only reinforces established knowledge but also uncovers new insights into how financial well-being impacts diabetes risk. These revelations underscore the necessity for holistic approaches in diabetes prevention and management strategies that consider both medical and socio-economic interventions. As we continue to unravel these complex relationships, our research paves the way for more targeted public health initiatives and underscores the importance of addressing income disparities as part of comprehensive diabetes care strategies.

# References

"Scikit-learn developers. 'Selecting Features Based on Importance Weights.' Scikit-learn: Machine Learning in Python. Accessed [Date you accessed the site]. https://scikit-learn.org/stable/auto_examples/feature_selection/plot_select_from_model_diabetes.html."

Cheruku, Ramalingaswamy, Damodar Reddy Edla, and Venkatanareshbabu Kuppili. 2020. Soft Computing Techniques for Type-2 Diabetes Data Classification. United States: CRC Press.

Liu, Huan, and Hiroshi Motoda. 2012. Feature Selection for Knowledge Discovery and Data Mining. Switzerland: Springer US.

Chen, Soy, Danielle Bergman, Kelly Miller, Allison Kavanagh, John Frownfelter, and John Showalter. "Using applied machine learning to predict healthcare utilization based on socioeconomic determinants of care." *Am J Manag Care* 26, no. 01 (2020): 26-31.

Panesar, Arjun. Machine Learning and AI for Healthcare: Big Data for Improved Health Outcomes. Germany: Apress, 2019.

Machine Learning for Healthcare Applications. United Kingdom: Wiley, 2021.

Computational Methods of Feature Selection. United States: CRC Press, 2007.

Gupta, Ananya. Feature Selection Techniques for Classification and Clustering. India: Vikatan Publishing Solutions, 2023.

Kuhn, Max., Johnson, Kjell. Feature Engineering and Selection: A Practical Approach for Predictive Models. United States: CRC Press, 2019.

Diabetes Detailed EDA with Conclusion." Kaggle, 2022. Accessed 2022 https://www.kaggle.com/code/bharat04/diabetes-detailed-eda-with-conclusion.

Horestani, Fariba Jafari, and Ginger Schwarz. "Survival Analysis of Young Triple-Negative Breast Cancer Patients." *arXiv preprint arXiv:2401.08712* (2024).

"Kaggle." Accessed February 12, 2024. https://www.kaggle.com/.

Schwarz, Ginger, and Fariba Jafari Horestani. "Prediction of Breast Cancer Recurrence With Machine Learning." In *Encyclopedia of Information Science and Technology, Sixth Edition*, pp. 1-33. IGI Global, 2025.

American Heart Association. "Diabetes Risk Factors." Last modified 2021. Accessed February 12, 2024. https://www.heart.org/en/health-topics/diabetes/diabetes-complications-and-risks/diabetes-risk-factors.

Kenney, J. 2024. "BMI and Diabetes: The Relationship Between Type 2 and Weight." diaTribe. Accessed February 12, 2024. https://diatribe.org/BMI-and-diabetes.

Horestani, Fariba Jafari, and Ginger Schwarz. "Survival Analysis of Young Triple-Negative Breast Cancer Patients." *arXiv preprint arXiv:2401.08712* (2024).

Zou, Quan, Kaiyang Qu, Yamei Luo, Dehui Yin, Ying Ju, and Hua Tang. "Predicting diabetes mellitus with machine learning techniques." *Frontiers in genetics* 9 (2018): 515.

Hasan, Md Kamrul, Md Ashraful Alam, Dola Das, Eklas Hossain, and Mahmudul Hasan. "Diabetes prediction using ensembling of different machine learning classifiers." *IEEE Access* 8 (2020): 76516-76531.

Khanam, Jobeda Jamal, and Simon Y. Foo. "A comparison of machine learning algorithms for diabetes prediction." *Ict Express* 7, no. 4 (2021): 432-439.

Shivadekar, Samit, Ketan Shahapure, Shivam Vibhute, and Ashley Dunn. "Evaluation of Machine Learning Methods for Predicting Heart Failure Readmissions: A Comparative Analysis." *International Journal of Intelligent Systems and Applications in Engineering* 12, no. 6s (2024): 694-699.

Shishehbori, Farnoush, and Zainab Awan. "Enhancing Cardiovascular Disease Risk Prediction with Machine Learning Models." *arXiv preprint arXiv:2401.17328* (2024).

Kee, Ooi Ting, Harmiza Harun, Norlaila Mustafa, Nor Azian Abdul Murad, Siok Fong Chin, Rosmina Jaafar, and Noraidatulakma Abdullah. "Cardiovascular complications in a diabetes prediction model using machine learning: a systematic review." *Cardiovascular Diabetology* 22, no. 1 (2023): 13.